\documentclass[letterpaper, 10 pt, conference]{ieeeconf}  % Comment this line out if you need a4paper

\usepackage{cite}
\usepackage{placeins}  
\usepackage{placeins} 
\usepackage{caption}
\usepackage{graphicx}
\usepackage{subcaption}
\usepackage{footnote}
\usepackage{balance}
\usepackage{authblk}
\usepackage{amsmath}   
\newcommand{\mathdagger}{^\text{\textdagger}}
\usepackage{amsfonts}
\usepackage{hyperref}
\usepackage{amssymb}
\usepackage{algorithm}
\usepackage{algpseudocode}
\usepackage{xcolor}
\definecolor{darkgreen}{rgb}{0, 0.5, 0}
\algnewcommand{\CppComment}[1]{\State \textcolor{darkgreen}{// #1}}
\usepackage{pgfplots}
\usepackage{hyperref}
\pgfplotsset{compat=1.17}
\usepackage{pgfplotstable}
\usepackage{booktabs}
\DeclareMathOperator*{\argmax}{arg\,max}
\IEEEoverridecommandlockouts                              % This command is only needed if 
\title{\LARGE \bf
AnyBipe: An End-to-End Framework for Training and Deploying Bipedal Robots Guided by Large Language Models
}

\author{Yifei Yao$^{1}$, Wentao He$^{2\mathdagger}$, Chenyu Gu$^{1\mathdagger}$, Jiaheng Du$^{1\mathdagger}$, Fuwei Tan$^{1}$, Zhen Zhu$^{1}$, and Junguo Lu$^{1,*}$

\thanks{$^{1}$ All authors are with Machine Vision and Autonomous System Laboratory, Department of Automation, School of Electrical Information and Electronic Engineering, Shanghai Jiao Tong University, Shanghai, China,  with the Key Laboratory of System Control and Information Processing, Ministry of Education of China, and with Shanghai Engineering Research Center of Intelligent Control and Management, Shanghai 200240, China. {\tt\small Email: $\{$godchaser, jiahengdu, wwddv1995, dyzz0928 , jglu $\}$@sjtu.edu.cn}, C. Gu: {\tt\small guchenyuFG@outlook.com}}%

\thanks{$^{2}$W. He is with University of Michigan - Shanghai Jiao Tong University Joint Institute , Shanghai Jiao Tong University, Shanghai, China {\tt\small Email: goodmorning\_hwt@sjtu.edu.cn}}%

\thanks{$\mathdagger$ The second to the fourth author contributed equally.}% <-this % stops a space

\thanks{$^*$ is for corresponding authorship.}%
}  % <-this % stops a space

\begin{document}

\maketitle 
\thispagestyle{empty}
\pagestyle{empty}

%%%%%%%%%%%%%%%%%%%%%%%%%%%%%%%%%%%%%%%%%%%%%%%%%%%%%%%%%%%%%%%%%%%%%%%%%%%%%%%%
\begin{abstract}

Training and deploying reinforcement learning (RL) policies for robots, especially in accomplishing specific tasks, presents substantial challenges. Recent advancements have explored diverse reward function designs, training techniques, simulation-to-reality (sim-to-real) transfers, and performance analysis methodologies, yet these still require significant human intervention. This paper introduces an end-to-end framework for training and deploying RL policies, guided by Large Language Models (LLMs), and evaluates its effectiveness on bipedal robots. The framework consists of three interconnected modules: an LLM-guided reward function design module, an RL training module leveraging prior work, and a sim-to-real homomorphic evaluation module. This design significantly reduces the need for human input by utilizing only essential simulation and deployment platforms, with the option to incorporate human-engineered strategies and historical data. We detail the construction of these modules, their advantages over traditional approaches, and demonstrate the framework's capability to autonomously develop and refine controlling strategies for bipedal robot locomotion, showcasing its potential to operate independently of human intervention.

\end{abstract}

%%%%%%%%%%%%%%%%%%%%%%%%%%%%%%%%%%%%%%%%%%%%%%%%%%%%%%%%%%%%%%%%%%%%%%%%%%%%%%%%
\section{Introduction}
With the integration of advanced control algorithms, enhanced physical simulations, and improved computational power, robotics has made significant strides\cite{9772943,10610566}. These innovations enable robots to perform tasks ranging from industrial automation to personal assistance with unprecedented efficiency and autonomy\cite{sutinys2022industrial, springer2021collaborative}. As industrial robotics matures, researches focus on humanoid robots, particularly in replicating human-like characteristics and enabling robots to perform human-oriented tasks\cite{ciardo2022jointtask}. Correspondingly, bipedal robots can be used to simulate the morphology of human lower limbs, thus providing a method to explore the locomotion skills of humanoid robots \cite{10611685, li2024reinforcement}.

Control strategies for bipedal robots typically leverage either traditional control methods or reinforcement learning (RL) methods\cite{robotics12010012,10125088}. Traditional approaches rely on problem abstraction, modeling, and detailed planning, while RL employs reward functions to iteratively guide robots toward task completion. Through repeated interactions with the environment, RL enables robots to refine control strategies and acquire essential skills, particularly excelling in trial-and-error learning in simulated environments, where robots adapt to complex terrains and disturbances.

Despite these advancements, training and deploying RL algorithms remains challenging. Effective reward design requires careful consideration of task-specific goals and the incorporation of safety constraints for real-world applications\cite{Toro_Icarte_2022,ernst2024introduction}. This complexity demands significant engineering effort in training, testing, and iterative refinement. Although reward shaping and safe RL offer potential solutions\cite{annurev:/content/journals/10.1146/annurev-control-042920-020211,HSU2023103811}, they often rely on prior experience, complicating the reward design process. Furthermore, bridging the gap between simulations and real-world conditions—the ``Sim-to-Real" challenge—remains difficult\cite{9606868}. Techniques such as \textit{domain randomization}\cite{slaoui2020robust}, which randomizes physical parameters to enhance agent robustness, and \textit{observation design}, which facilitates task transfers across varied terrains, prove formidable, but still require real-world testing and human feedback for model selection and fine-tuning\cite{li2024reinforcement, gu2024humanoid,kwon2023reward}.

The integration of large language models (LLMs) into robotics offers the prospect of reducing human work in such process. Known for their capabilities in code generation and missing planning, LLMs are being widely applied to complex robotics applications\cite{NEURIPS2022_67496dfa}. For instance, they play a pivotal role in embodied intelligence by enabling the dynamic creation of action tasks\cite{10610744}. Recent developments have explored the utility of LLMs in improving reward function design and refining from simulation performance—key areas that reduce the need for human intervention\cite{ma2024dreureka}. However, a comprehensive framework that designs suitable policies and automatically implements properly trained models in real-world remains lacking. To address this issue, we propose a novel framework that leverages LLMs to optimize the entire training-to-deployment process. The framework minimizes human involvement, enables autonomous design, training, and deployment of RL algorithms, and supports exploring strategies from scratch or enhancing existing ones.

\begin{figure*}[htbp]     
    \centering
    \includegraphics[width=0.95\textwidth]{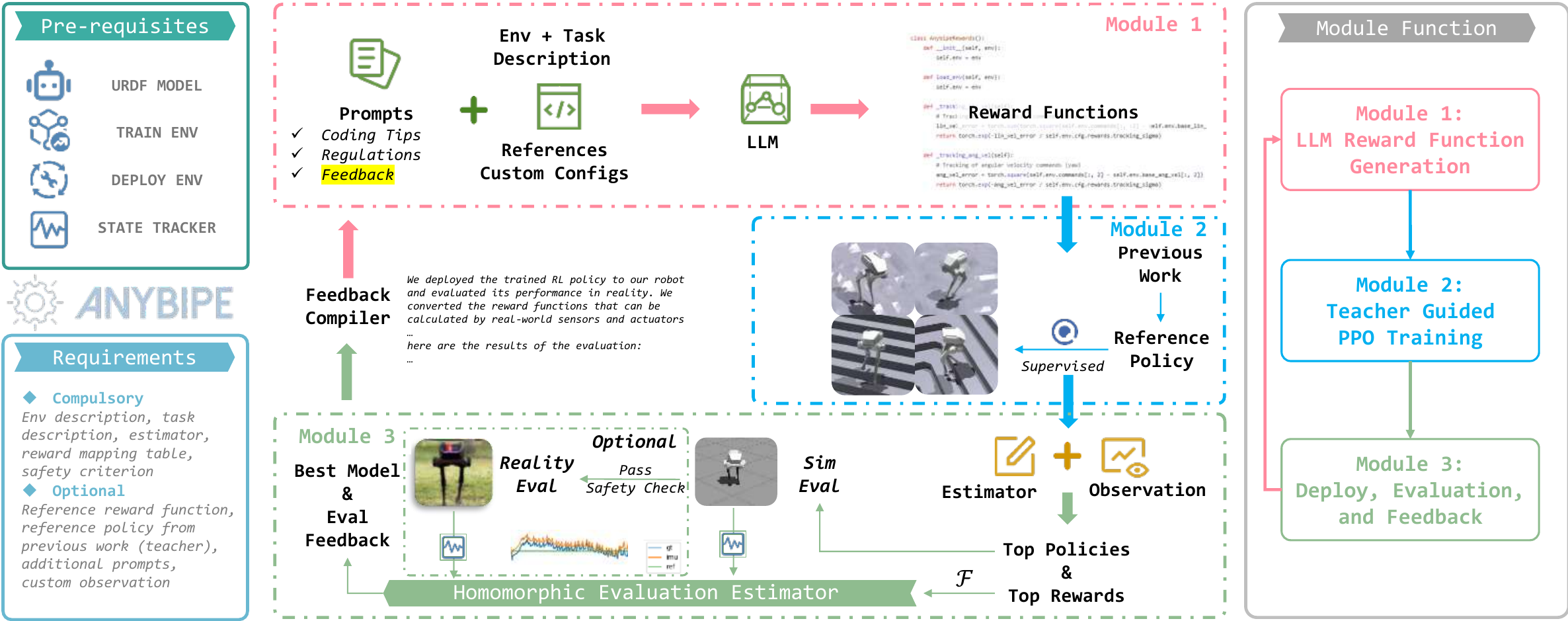}  
    \caption{Our frameworks are organized in three interconnected modules. After receiving all pre-requisites and requirements, the framework generates reward function via LLM, trains it in simulation and evaluates in both simulation and reality, providing important feedback. The whole procedure requires minimum human labor. } 
    \label{fig:your-label}
\end{figure*}

We present \textit{AnyBipe}\footnote{\tt\footnotesize\url{https://github.com/sjtu-mvasl-robotics/AnyBipe.git}}, the \textbf{first} end-to-end framework for training, simulating, and deploying bipedal robots, guided by LLMs. This framework consists of three interconnected modules and embodies a closed-loop process for continual refinement and deployment. Our main contribution are listed as follows:

\begin{enumerate}
\item \textbf{Minimized Human Intervention}: The proposed framework operates autonomously, requiring only minimal initial user input to oversee the entire process from training to deployment. No additional human intervention is needed during the workflow.

\item \textbf{LLM-Guided Reward Function Design}: Leveraging large language models, the framework generates suitable reward functions from predefined prompts. Through iterative refinement, it allows users to design customized RL reward functions from the ground up.

\item \textbf{Incorporation of Prior Knowledge in Training}: The framework enables the integration of pre-trained models and conventional control algorithms, which enhances RL training stability and facilitates the migration of traditional control implementations into the proposed system.

\item \textbf{Real-World Feedback Integration via Homomorphic Evaluation}: This module converts real-world sensor and actuator feedback into formats compatible with simulation, enabling LLMs to bridge the gap between the training environment and real-world deployment. As a result, it allows for adaptive adjustments to the reward function based on actual feedback.

\end{enumerate}

Experimental trials on bipedal robots traversing both flat and complex terrains have shown that \textit{AnyBipe} can significantly enhance real-world deployment performance. Compared to manually designed reward functions, those generated by \textit{AnyBipe} lead to faster convergence and more stable training outcomes, also preventing falling into high-reward but low-usability solutions. Furthermore, homomorphic real-world feedback allows LLMs to understand the impact of the reward function on actual deployment performance, thereby improving the corresponding design and improving Sim-to-Real performance.

This paper is organized as follows: First, we introduce the design principles and implementation methods of the framework modules in detail. We then describe the experiments used to verify the role of each module and express the training and deployment effects of the framework. Finally, we concluded our contributions and propose future research directions for our framework.

\section{Related Works}

\textbf{Reinforcement Learning for Bipedal Robots.}
Reinforcement learning (RL) has achieved significant success in enabling legged robots to learn locomotion and motion control through data-driven methods, allowing them to adapt to diverse environmental challenges\cite{peng2018deepmimic, hwangbo2019learning, rudin2022learning, caluwaerts2019data, kalakrishnan2011learning, lee2020learning}. Research has introduced RL-based locomotion strategies, training in simulation environments like MuJoCo\cite{todorov2012mujoco} and Isaac Gym\cite{makoviychuk2021isaac}. Recent studies have shifted these techniques, originally designed for quadruped robots and robot arms, to bipedal robots like Cassie\cite{apgar2018fast,9561705}. Additional approaches explore imitation learning\cite{wu2024infer}, motion planning\cite{9981903}, and robust RL strategies\cite{li2024reinforcement}, enabling bipedal robots to perform tasks like running, stair climbing, and complex maneuvers\cite{siekmann2021sim, siekmann2021blind, haarnoja2024learning}. Building on these advancements, our work utilizes the Isaac Gym environment and proposes a semi-supervised RL framework to transfer the original work and effectively improve the learning performance.

\textbf{Large Language Model Guided Robotics.}
Large language models (LLMs) have demonstrated considerable capabilities in task understanding\cite{huang2022inner, huang2022plan4grounded}, semantic planning\cite{ahn2022saycan, zhang2023bootstrap}, and code generation\cite{wang2021code, jiang2021cure, liang2023code}, allowing them to be effectively integrated into a variety of robotic tasks. LLMs automate environmental analysis\cite{wang2023gensim, wang2023robogen}, reward functions design\cite{10, yu2023language}, and task-action mapping\cite{driess2023palm, vemprala2024chatgpt}. However, challenges such as data scarcity, real-time performance, and real-world integration remain\cite{zeng2023large}. In addition, due to the lack of awareness of actual data, LLM-driven code design sometimes needs to be optimized through human feedback. Our framework addresses the existing limitations by adopting environmental characteristics and safety constraints as priors, and uniquely combines homomorphic feedback from real-world applications, reducing the need for human feedback in the process.

\textbf{Sim-to-real Training and Deploying Techniques.}
The gap between simulated environments and real-world conditions, known as the ``reality gap'', presents significant challenges for deploying RL strategies in robotics\cite{brunnbauer2022latent, betz2023bypassing}. Techniques such as domain randomization\cite{muratore2022robot, peng2018sim, tiboni2023domain} and system identification\cite{aastrom1971system, deisenroth2011pilco, tan2018sim} are widely used to address this issue. Researchers have proposed sim-to-real solutions for bipedal robots to handle tasks such as turning and walking\cite{9812015,10000225}. Recent work has also integrated LLMs to enhance environmental modeling and reward function design, making simulations more reflective of real-world complexity\cite{ma2024dreureka}. However, most approaches still rely on separate training in simulation and real-world evaluation, often using human feedback to assess sim-to-real effectiveness. Our work extends these techniques by introducing an evaluation loop that continuously monitors sim-to-real performance during deployment.

% \textbf{Robot State Estimation and Evaluation.}
% Accurate robot state estimation is crucial for effective control in both simulated and real-world environments. Traditional methods, such as the extended Kalman filter (EKF)\cite{doi:10.1177/0278364919894385}, typically fuse IMU and robot joint state data to estimate pose and velocity, ensuring stability during tasks like locomotion \cite{ BAI2023100909, yu2020adaptive}. However, handling simulation and real-world feedback separately can limit performance during deployment \cite{9158349}. To address this issue, we integrate IMU and robot joint state data from both the Gazebo simulation and real-world trials\cite{9714001}, generating time-series plots for key variables such as pose and velocity. These data are fed to a large language model (LLM), which adjusts and optimizes the control algorithms based on the feedback.
% The LLM iteratively refines the control algorithms, improving the robot's ability to track desired states in subsequent training cycles. This feedback loop continuously reduces the sim-to-real gap, enhancing state estimation accuracy and robustness with each iteration, ultimately improving real-world performance.

\section{Methods}

In this section, we introduce the \textit{AnyBipe} framework in detail. The framework consists of three modules that collectively aim to enhance reward design, facilitate semi-supervised RL training, and automate evaluation and feedback. Users are required to prepare the URDF model, task description, basic RL environment, and an operational control platform such as ROS\cite{quigley2009ros}. A runtime data collector, known as the State Tracker, which gathers essential runtime data such as IMU data, joint positions, velocities, and torques, is also essential. Starting with a well-engineered prompt framework for code generation, we enable LLMs to write appropriate reward functions based on user-provided patterns, while considering their functionality and potential safety issues. The generated reward functions are then trained in modified RL environments, aided by cold start techniques to improve convergence. The top-performing policies are subsequently validated using a homomorphic estimator to assess their safety and compare their performance with the training stage. All estimation results are then compiled into feedback prompts for the next iteration's improvement. The procedural steps are encapsulated in Algorithm \autoref{algorithm:1}.

\begin{algorithm}[ht]
\caption{AnyBipe Framework Process}
\label{algorithm:1}
\begin{algorithmic}[1]

\State \textbf{Pre-requisites}: URDF model $O$, training environment $\mathcal{T}$, deployment environment $\mathcal{R}$, robot state tracker $st$

\State \textbf{Require}: Environment description $\mathcal{D}(\mathcal{T})$, prompt set $p$, training environment estimator $\mathfrak{E}_{train}$, homomorphic estimator mapping function $\mathcal{F}$, safety evaluation criterion $SA$. RL algorithm $\operatorname{RL}$, LLM model $\operatorname{LLM}$, feedback prompt compiler $\operatorname{COMPILE}$

\State \textbf{Optional}: Human-engineered reward function $\mathbf{R}_{ref}$, reference policy (previous work) $\pi_{ref}$, additional prompts $p_{add}$, critical human factor to be observed $obs_{c}$

\State \textbf{Hyperparameters}: Iteration $N$, number of reward candidates $K$, best sample percentage $c_{bs}$, teacher model coefficient $\beta$, environment estimator coefficient $c_{e}$, human factor observation coefficient $c_{obs}$ 
 
\State \textbf{Input}: Task description $\mathcal{D}$

\State $\mathbf{R}_{ref} \gets$ \textbf{if} pre-defined \textbf{then} reference reward \textbf{else} None
\State $\pi_{ref} \gets \operatorname{RL}(\mathbf{R}_{ref})$ or user-defined \textbf{if} exists \textbf{else} None

\For{$i \gets 1$ \textbf{to} $N$}
    \CppComment {Module 1: Reward Function Generation}
    \State $p_{in} \gets p + p_{add} + p_{feedback}$
    \State $\mathbf{R} \gets \operatorname{LLM}(\mathcal{D}, \mathcal{D}(\mathcal{T}), p_{in}, \mathbf{R}_{ref})$

    \CppComment {Module 2: Semi-Supervised RL Training}
    \State $\Pi, \operatorname{Obs} \gets \operatorname{RL}(\mathcal{T}, O, \mathbf{R}, \pi_{ref}, \beta)$
    \CppComment {Module 3: Automated Evaluation and Feedback}
    \State $n_{bs} \gets \lceil c_{bs} \cdot K \rceil$
    \State $p_{feedback} \gets$ None
    \State $Criterion \gets c_{e} \cdot \mathfrak{E}_{train}(\Pi) + \sum c_{obs} \cdot \operatorname{Obs}$
    \State $\mathbf{R}_{bs}, \pi_{bs} \gets \argmax_{Criterion}(\mathbf{R}, \Pi)$
    %\CppComment {Mapping rewards to real world measurements}
    \State $\hat{\mathbf{R}}_{bs} \gets \mathcal{F}(\mathbf{R}_{bs})$

    \ForAll{$\pi, \hat{\mathbf{R}}$ \textbf{in} $\pi_{bs}, \hat{\mathbf{R}}_{bs}$}
        \State $\pi_{real} \gets (\mathcal{T} \to \mathcal{R})(\pi)$
        \State $\mathfrak{E}_{sim} \gets \operatorname{EVAL}_{sim}(\hat{\mathbf{R}}(st(O)), \pi_{real})$
        \State $p_{feedback} += \operatorname{COMPILE}(p, \mathfrak{E}_{sim})$

        \If{$SA(\mathfrak{E}_{sim})$ \textbf{is} True}
            \State $\mathfrak{E}_{real} \gets \operatorname{EVAL}_{real}(\hat{\mathbf{R}}(st(O)), \pi_{real})$
            \State $p_{feedback} += \operatorname{COMPILE}(p, \mathfrak{E}_{real})$
        \EndIf
    \EndFor

    \State $\mathbf{R}_{ref}, \pi_{ref} \gets \argmax_{Criterion}(\mathbf{R}_{bs}, \pi_{bs})$
    \State $p_{feedback} += \operatorname{COMPILE}(p, \pi_{ref})$
\EndFor

\State \textbf{Output}: Best policy $\pi$, best reward function $\mathbf{R}$
\end{algorithmic}
\end{algorithm}

\subsection{Module 1: LLM Guided Reward Function Design with Proper Prompt Engineering}
\label{sec:3.1}

We refer to the reward function design in the Eureka algorithm \cite{10}, which enables LLMs to autonomously improve and iterate reward functions with predefined environmental and task descriptions $\mathcal{D}(\mathcal{T})$. However, initial usability issues require multiple iterations for viable training code. Furthermore, Eureka often overlooks discrepancies between training $\mathcal{T}$ and real environments $\mathcal{R}$, resulting in computationally expensive but minimally effective reward functions that may induce undefined behaviors \cite{kim2024not}. The framework also lacks comprehensive safety considerations for tasks such as bipedal movement, despite attempts to integrate safety through Reward-Aware Physical Priors (RAPP) and LLM-led Domain Randomization \cite{ma2024dreureka}.

To address these challenges, we have developed a robust context establishment mechanism to assist LLMs in constructing reliable and secure reward functions. As outlined in the algorithm, we employ an iterative ReAct (Reason + Act) framework \cite{yao2022react} to improve the coding results based on real-world performance. The prompts are structured into three distinct phases: the system prompt stage, the task specification (initial user) stage, and the feedback (user) stage. The system prompt stage is dedicated to generating knowledge prompts \cite{liu2021generated} for the RL environment, as well as utilizing few-shot prompting \cite{brown2020language} for reward function generation. Relevant system variables, such as ``root states," ``feet height," and ``rigid body state," are extracted from the environment and cataloged into a reference table to prevent the LLM from inventing non-existent variables. This approach also improves the integration of environmental variables within the set $\mathcal{D}(\mathcal{T})$. For more intricate tasks, such as humanoid walking with a desired pattern, we encourage users to incorporate custom prompts following our ``coding references" guide. This enables LLMs to learn from human-engineered reward functions and autonomously generate task-specific code. Additionally, a section on ``coding restrictions" is provided as negative prompts, urging LLMs to account for safety constraints such as contact forces, degrees of freedom (DoF) limits, and torque limitations. Our experimental findings confirm that LLMs can effectively incorporate these safety restrictions and environmental variables, thereby designing highly efficient reward functions. These results underscore the models' outstanding ability to track contextual information and adhere to directives \cite{ouyang2022training, wei2023larger}.

\begin{figure}[htbp]
    \centering
    \includegraphics[width = 0.49\textwidth]{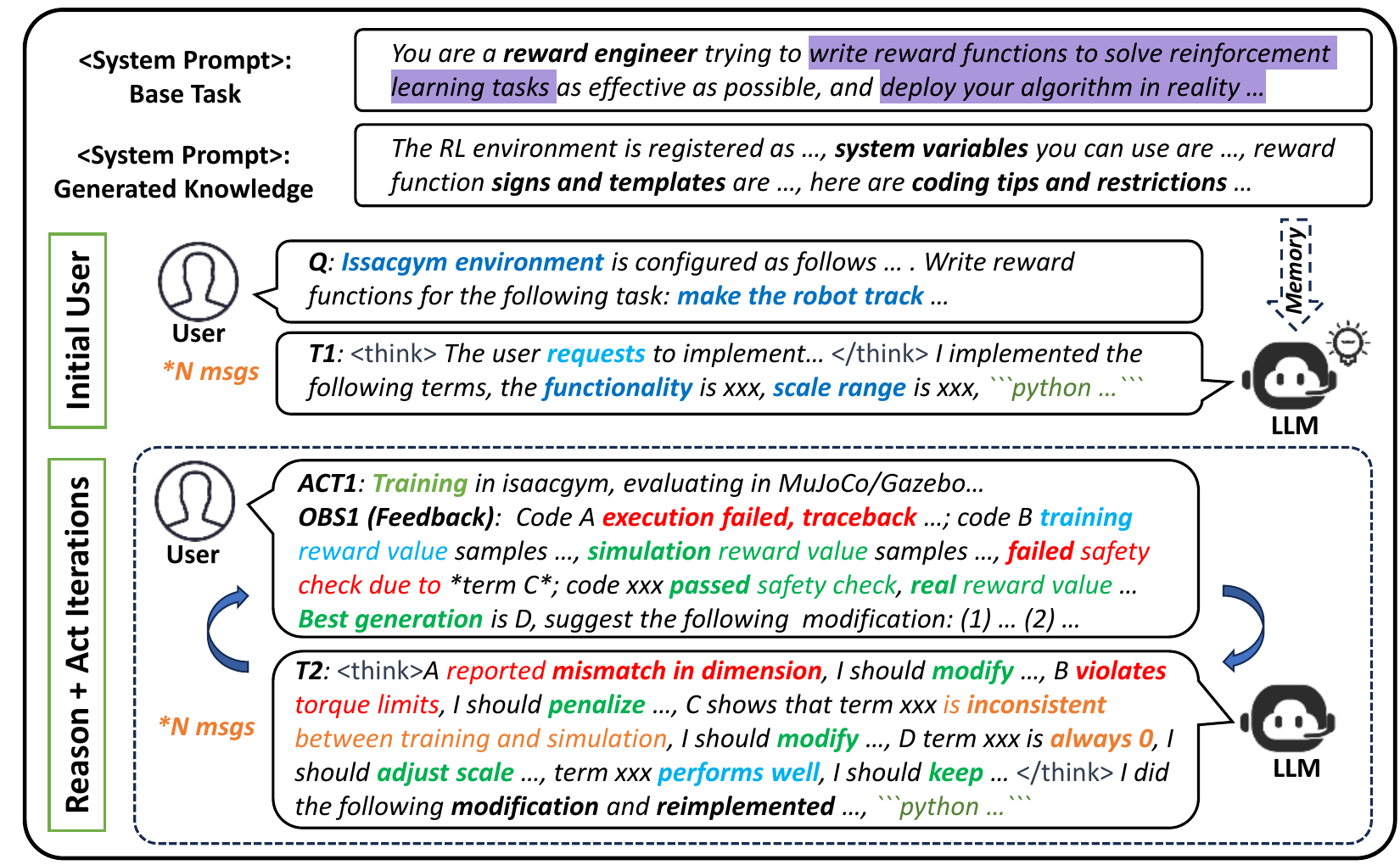}
    \caption{Demonstration of LLM reward generation iterations. LLMs are encouraged to adopt CoT form output and adjust its generation results according to feedback. Each iteration requires generating $N= K$ reward function samples, they are separately trained and evaluated, and the best performance one is set as template for next LLM generation.}
    \label{fig:llm function}
\end{figure}

The initial task-specific stage serves to provide essential information for training, including a description of the robot, key parameters such as base height and average foot air time, as well as an introduction to reward-writing techniques. Following the first iteration, this stage is replaced by a comprehensive reward function evaluation scheme, detailed in \autoref{sec:3.3}, which functions as a feedback prompt. This prompt incorporates multiple components, including the performance of the reward function, the consistency of the homomorphic estimator with the reward functions, and whether the policy adheres to safety restrictions. The evaluator integrates pre-defined human preferences, such as human factor observation (\(obs_c\)) and user-defined observation terms labeled with ``gt" to maintain consistency across LLM-generated reward functions. The success rate is also considered to identify the best-performing sample, which is then fed back to guide the LLM in adjusting terms and coefficients.

Additionally, we incorporate self-consistency techniques \cite{wang2022self} to assist the LLM in assessing the performance of individual reward terms, determining whether they exhibit good or poor performance, and making necessary modifications. In cases where the code fails to execute, error logs related to the reward file are extracted and provided as feedback. This stage also offers adjustment tips to further refine the reward function. We also guide the LLMs to generate reward functions in the form of Chain-of-Thought (CoT) \cite{wei2022chain}, providing a step-by-step explanation of their implementation. The generated code is instructed to be presented within a code block at the end of the output, which is subsequently processed by our script to remove the $<think>$ tags and extract the target code within the code block. Our experimental results demonstrate that utilizing the CoT approach enhances the LLM's ability to accurately identify and address issues within the code implementation.

\subsection{Module 2: RL Training Adopting Reference Policy for Cold-Start}

In the initial phase of training, where Large Language Models (LLMs) generate rewards without external feedback, the reward terms often exhibit poor scaling, which can result in slow convergence—a challenge commonly referred to as the ``cold start" problem. To address this issue, the integration of a reference policy can provide guidance, ensuring that the policy network converges toward the desired objective. This section outlines the adaptations made to direct reinforcement learning (RL) training toward the target actions, utilizing frameworks from Legged Gym and leveraging the Proximal Policy Optimization (PPO) algorithm \cite{Schulman2017ProximalPO} as the foundational approach. We assume the existence of a baseline policy, denoted as $\pi_{\text{ref}}$, which is derived from pre-existing functional policies. Interfaces are provided to accommodate three types of teacher inputs: policies implemented using traditional control techniques such as Model Predictive Control (MPC) or Whole Body Control (WBC), previously trained policies in formats like PyTorch or ONNX, and simple locomotion distributions, such as sine waves.
% , operating with inputs $s_{\text{ref}}$ that are a subset of the inputs used by the training RL algorithm, $s_{\text{ref}} \subset s$. Thus, if

% \begin{equation}
% a_{t} = \pi(s_t, a_{t-1})
% \end{equation}

% is satisfied by our trained policy $\pi$, the reference policy should also satisfy

% \begin{equation}
% a_{t,\text{ref}} = \pi_{\text{ref}}(s_t, a_{t-1,\text{ref}})
% \end{equation}

To enhance the PPO algorithm, we modify the objective function given below:

\begin{equation}
\begin{aligned}
    L^{\text{CLIP}}(\theta) = & \hat{\mathbb{E}}_t [\min(r_t(\theta) \hat{A}_t, \text{clip}(r_t(\theta), 1 - \epsilon, 1 + \epsilon) \hat{A}_t) \\
    &+\beta \operatorname{KL}[\pi_{\text{ref}}(\cdot | s_t), \pi_{\theta}(\cdot | s_t)]],
\end{aligned}
\end{equation}

\noindent where $r_t(\theta) = \frac{\pi_\theta(a_t | s_t)}{\pi_{\theta_{old}}(a_t | s_t)}$ denotes the probability ratio, $\hat{A}_t$ is an estimator of the advantage at time $t$, $\epsilon$ is a small positive number, and $\beta$ is a coefficient that measures the divergence between the reference policy and PPO policy.

This integration enables control over the similarity between the trained policy and the reference policy. However, abstracting the reference policy $\pi_{\text{ref}}$ into the same probabilistic framework as the PPO policy presents challenges. Despite this, given the deterministic nature of actions $a_t$ for a specific state $s$ and previous action $a_{t-1}$, and assuming sufficient environmental observations, we can approximate the distribution $\pi_{\text{ref}}$ as a Dirac distribution, and the differences between $\pi$ and $\pi_{\text{ref}}$ can be described as follows:

\begin{equation}
\begin{aligned}
     \hat{\mathbb{E}}_t\left[\operatorname{KL}( \pi_{\text{ref}}, \pi_{\theta})\right] \approx 
     \frac{1}{N}\sum_{i=1}^N \left[\log(\sqrt{2\pi \sigma_{\theta,i}^2}) + \frac{(a_{ref} - \mu_{\theta,i})^2}{2\sigma_{\theta,i}^2}\right].
\end{aligned}
\end{equation}

With adequate observations, this approximation becomes reliable. The weight of this term should be set to a relatively small value in order to guide the policy net out of local minimum, but not preventing it from exploring better solution.

In the \textit{AnyBipe} framework, we have developed a template for deploying existing policies as teacher functions, illustrating how such policies can be integrated into the reinforcement learning (RL) training process. The integration mechanism operates as follows: if a human-engineered reward function is designated as the ground-truth reward, it is trained prior to the iteration, and the trained policy is subsequently set as the teacher. Alternatively, users can provide their own implemented policies following our template. After each iteration, the reference policy is updated to the trained policy with the best performance if it passes a safety check. This metric compares the performance of human-engineered policies with those generated by Large Language Models (LLMs), offering essential insights into the correctness of metric implementation. Furthermore, it safeguards against the degradation of the LLM-generated reward function by tracking how the behavior of reward function terms evolves after LLM modification, following the same Markov Decision Process (MDP) path.

\subsection{Module 3: Feedback From Simulation and Deployment Stage With Minimal Human Effort}
\label{sec:3.3}

Traditionally, Eureka-like algorithms only collect training feedback for the subsequent generation, with generated policies left to engineers for examination and deployment. In contrast, our framework automates the entire process using Python, Bash, and C++ scripts, establishing a robust safety check criterion and providing a numerical measure for the sim-to-real gap, referred to as the ``homomorphic estimator". This estimator enables the Large Language Model (LLM) to identify which terms have a significant impact on deployment procedures.

To establish a general criterion for evaluating policy performance, we define a set of observations with ``ground-truth" (gt) labels, which remain consistent across all generations. For all our experiments, this set includes linear velocity tracking results, angular velocity tracking results, and survival time. The product of these three indicators is considered the base factor, termed the \textit{environment estimator} $\mathfrak{E}_{train}$, which gauges the success of policy execution. Additionally, we provide users the flexibility to define custom observation/reward factors, termed human factors $obs_c$, within the configuration file, along with their corresponding weights $c*{obs}$. The weighted sum of these factors is then used to compute the total score, with policies ranked accordingly to select the best candidates for simulation and deployment validation.

\begin{equation}
\mathbf{R}_{bs}, \Pi_{bs} = \argmax_{n_{obs}} c_{e} \cdot \mathfrak{E}_{train}(\Pi) + \sum c_{obs} \cdot \operatorname{Obs}_c.
\end{equation}

\begin{figure}[htbp]
    \centering
    \includegraphics[width = 0.47\textwidth]{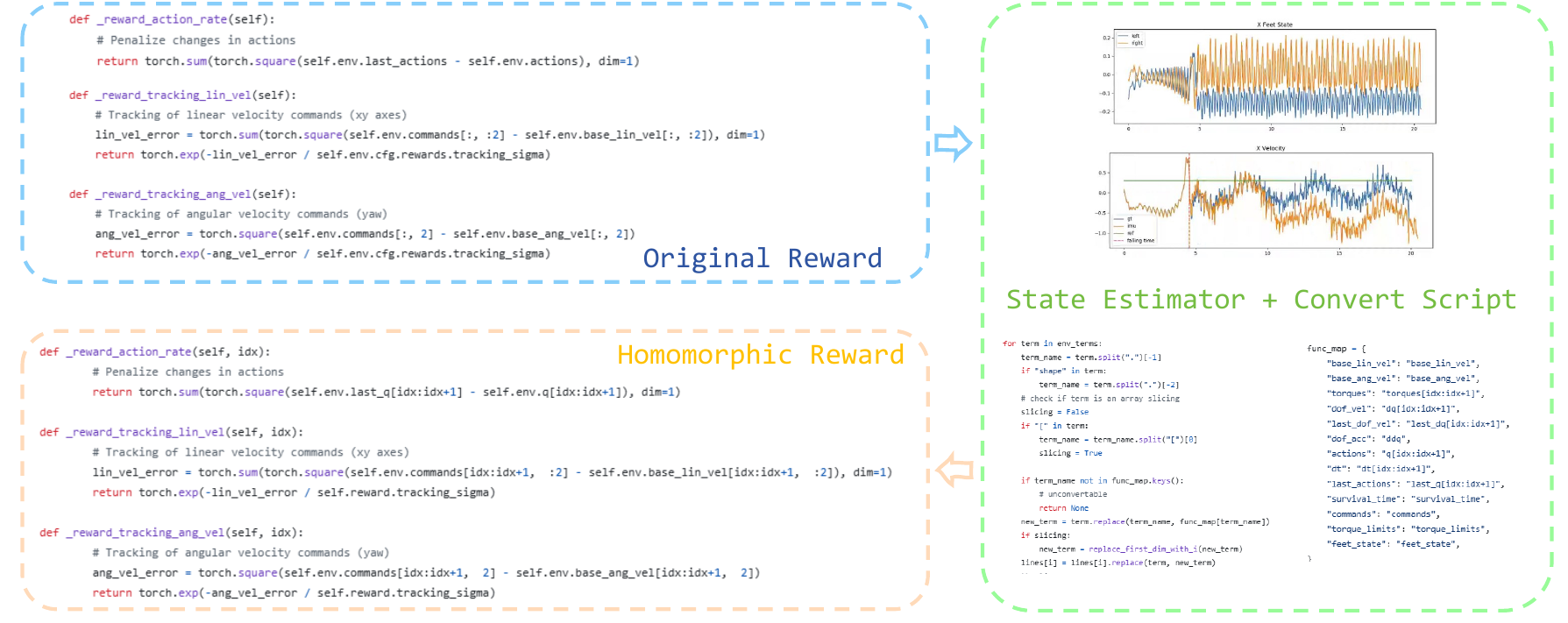}
    \caption{Homomorphic reward function conversion procedure.}
    \label{fig:homo}
\end{figure}

We automatically deploy the policy file, in either JIT or ONNX format, to a user-specified simulation environment such as Gazebo or MuJoCo. The simulation script runs for at least $num\_steps\_per\_env \times step\_dt$, as defined in the OnPolicyRunner under the rsl\_rl library, ensuring that the simulation duration matches or exceeds a full training epoch. Data from the robot, including IMU readings, fall-off time (or maximum if no fall-off occurs), joint speed, velocity, and torques, are collected for evaluation by our \textit{homomorphic evaluator}. Given that the training (\(\mathcal{T}\)) and real-world environments (\(\mathcal{R}\)) are isomorphic, we define a homomorphism \(\mathcal{F}: \mathcal{T} \to \mathcal{R}\), ensuring that the real-world evaluation metric \(\hat{\mathbf{R}} = \mathcal{F}(\mathbf{R})\) mirrors the reward function. Users are required to provide a mapping table linking IsaacGym system variables to simulation-tracked state variables (e.g., root\_state in IsaacGym to part of qpos in MuJoCo). Fig .\ref{fig:homo} illustrates this conversion procedure, where the reward functions are transformed into new estimators that maintain consistency between training and simulation. The evaluator then provides a quantitative measure of the sim-to-real gap, identifying specific reward terms that may require improvement, thus assisting the LLM in refining its approach.

To ensure the policy is safe for real-world deployment, selected policies must pass a safety check, denoted as $SA$, before being compiled by ROS. Users are required to implement $SA$ following the template provided. $SA$ evaluates simulation data, including torques, positions, ground contact forces, and other relevant factors. If all checks pass, $SA$ returns true, enabling the script to automatically compile and execute the reality deployment code (this stage can be passed if no reality robots are available). If any checks fail, the script identifies the failed terms and generates a feedback prompt. The reality evaluator operates similarly to the simulation evaluator, gathering and evaluating data. Users are allowed to monitor the process and can stop the deployment if dangerous behavior is detected. External stops are recorded by the script, indicating that the reality deployment has failed; otherwise, the deployment is considered successful. These steps also provide a score that contributes to policy selection, with the best generation calculated as follows:

\begin{equation}
\begin{aligned}
    &R_{best},\pi_{best} = \\&\argmax\left[ [\mathcal{F}(\mathbf{R}_{bs})_{sim} + \mathcal{F}(\mathbf{R}_{bs})_{real}](\mathcal{T} \to \mathcal{R})(\Pi_{bs})\right].
\end{aligned}
\end{equation}

Except for the final reality deployment stage, which is recommended to have a supervisor (though it can still operate without human oversight, or with VLM and cameras substituting human functions), the entire cycle operates autonomously. Evaluation, safety check, and deployment procedure are working out human-free. Even the initial reference rewards or pre-existing policies are optional, as LLMs demonstrate significant potential in handling zero-shot generation tasks \cite{kojima2022large}. This framework offers a method for automatically adjusting or tuning reward functions for robot engineers, while also providing a solution for developing RL policies from scratch for robots that have no prior RL implementations.

\section{Experiments}
\subsection{Experimental Setup}

\begin{figure}[htbp]
    \centering
    \includegraphics[width = 0.4\textwidth]{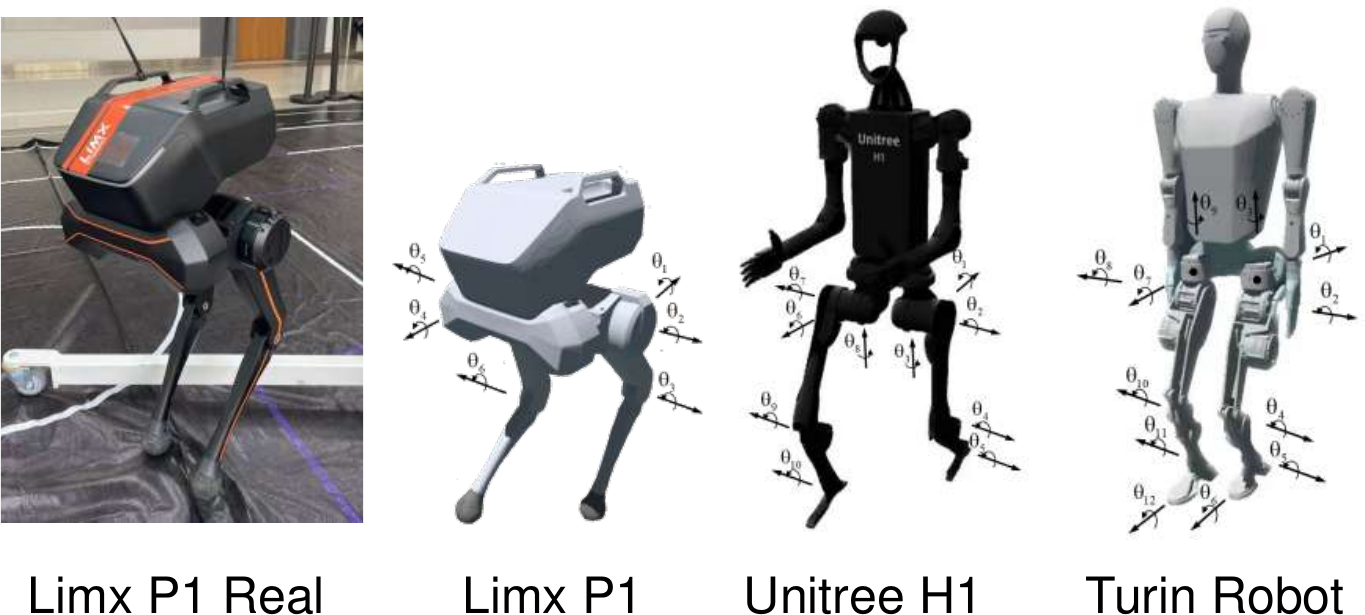}
    \caption{Robot and DOF definitions.}
    \label{fig:dof}
\end{figure}
Our experiments were conducted with three types of bipedal robots: the 6 DoF pointfoot robot P1 from Limx Dynamics, the 10 DoF humanoid Unitree H1 (lower body only), and the 29 DoF humanoid from Turin Robots (12 lower body DoF, with other DoF fixed during training), as shown in Fig. \ref{fig:dof}. Each robot was equipped with different initial prompt sets: P1 used human-engineered reward functions as the ground truth for comparison; Unitree H1 used some human-engineered functions as templates; and Turin, with no prior RL implementation, utilized templates partially adapted from the open-source humanoid-gym project, with no ground-truth reward or reference policy. LLMs used include OpenAI GPT-4o\cite{achiam2023gpt}, Anthropic Claude-3.5-Sonnet \cite{claude35}, and Deepseek-R1 \cite{guo2025deepseek}, to validate the versatility of our prompt groups and scripts. The framework was trained on a GPU server with 4 NVIDIA RTX 3090 GPUs. The robots were trained on common and trimesh terrains, with task settings including $N = 5$ iterations, $K = 16$ samples, and the top $15\%$ (3 best samples) selected for each task. Thanks to the implemented GPU allocator algorithm, we are able to complete a run with 80 trainings in total in 8 hours, including LLM generation and evaluation time. For 2 humanoids, the framework functions with only \textit{simulation} evaluation step with $SA$ feedback. For P1, we are able to perform a real-world deployment test to validate our auto-deployment process. The GPU server is hardwared to connect to robot controllers, and simulation environments (MuJoCo mjviewer and Gazebo) are set to function under headless mode.

\autoref{tab:1} outlines general \textit{environment estimators} in the form of reward functions. These estimators maintains the same form across all training and evaluating process.

\begin{table}[h]
\centering

\begin{tabular}{l l }
\toprule
\textbf{Reward Name}  & \textbf{Expression Form}\\
\midrule

Survival  & $ R_{surv}= \int_0^{t_{\text{term},i}} dt$  \\
Tracking Linear Velocity  & $R_{vel} = \exp(- \frac{\|v-v_{ref}\|^2}{\sigma_{l}^2})$  \\
Tracking Angular Velocity  & $R_{angl}=\exp(- \frac{\|\omega-\omega_{ref}\|^2}{\sigma_{a}^2})$   \\
Success  & $ R_{succ} = R_{surv} \cdot R_{vel} \cdot R_{angl}$ \\
\bottomrule

\end{tabular}
\caption{Examples of important rewards}
\label{tab:1}
\end{table}

\subsection{Module Analysis}
Before introducing the outcome for the whole framework process, we first conduct several experiments seperately on each module to examine its effectiveness. We evaluated the LLM performance, RL-training performance, homomorphic performance, and whether $SA$ criterion functions normally.

\begin{figure*}[htbp]
    \centering
    \begin{minipage}{0.3\textwidth}
        \centering
        \includegraphics[height=3.5cm, keepaspectratio]{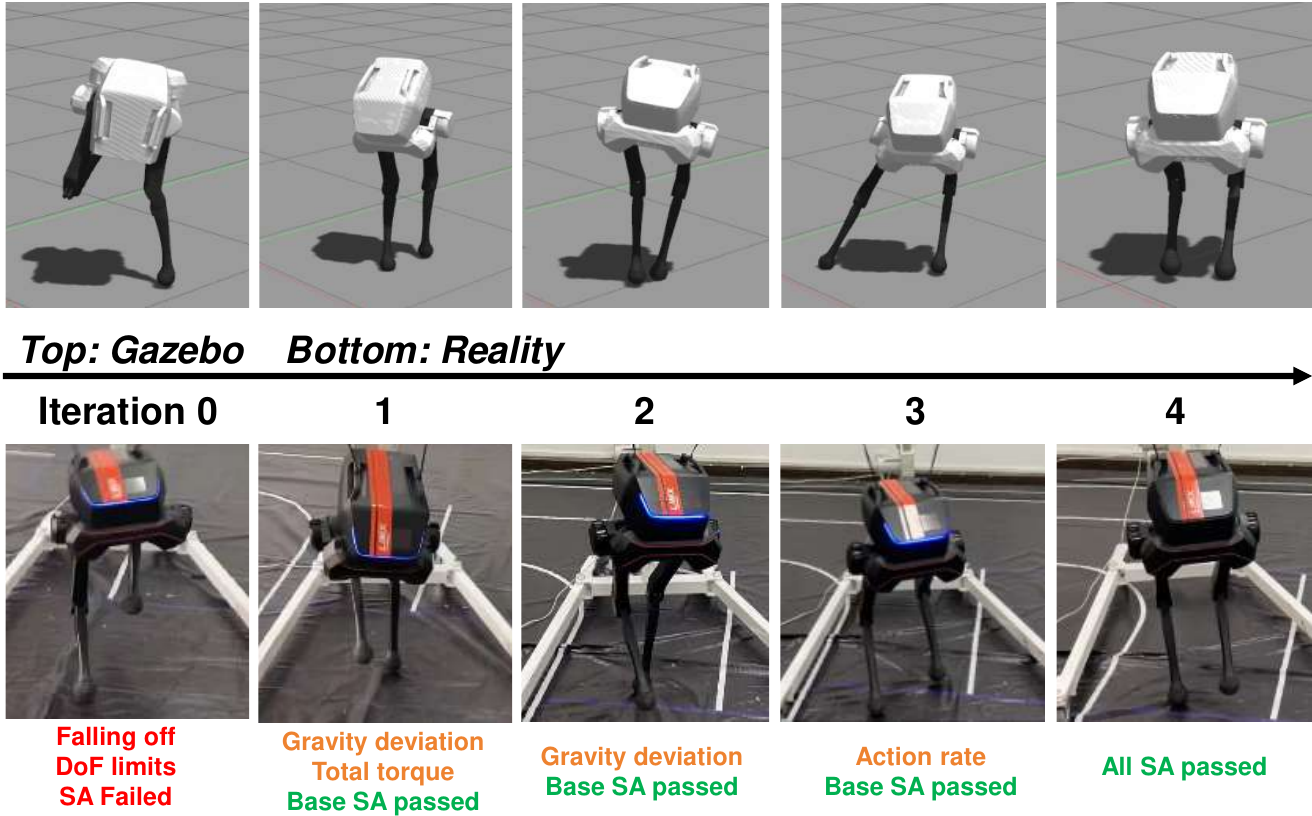}
        \caption*{P1 iterations.}
    \end{minipage}%
    \hspace{0.02\textwidth}
    \begin{minipage}{0.3\textwidth}
        \centering
        \includegraphics[height=3.5cm, keepaspectratio]{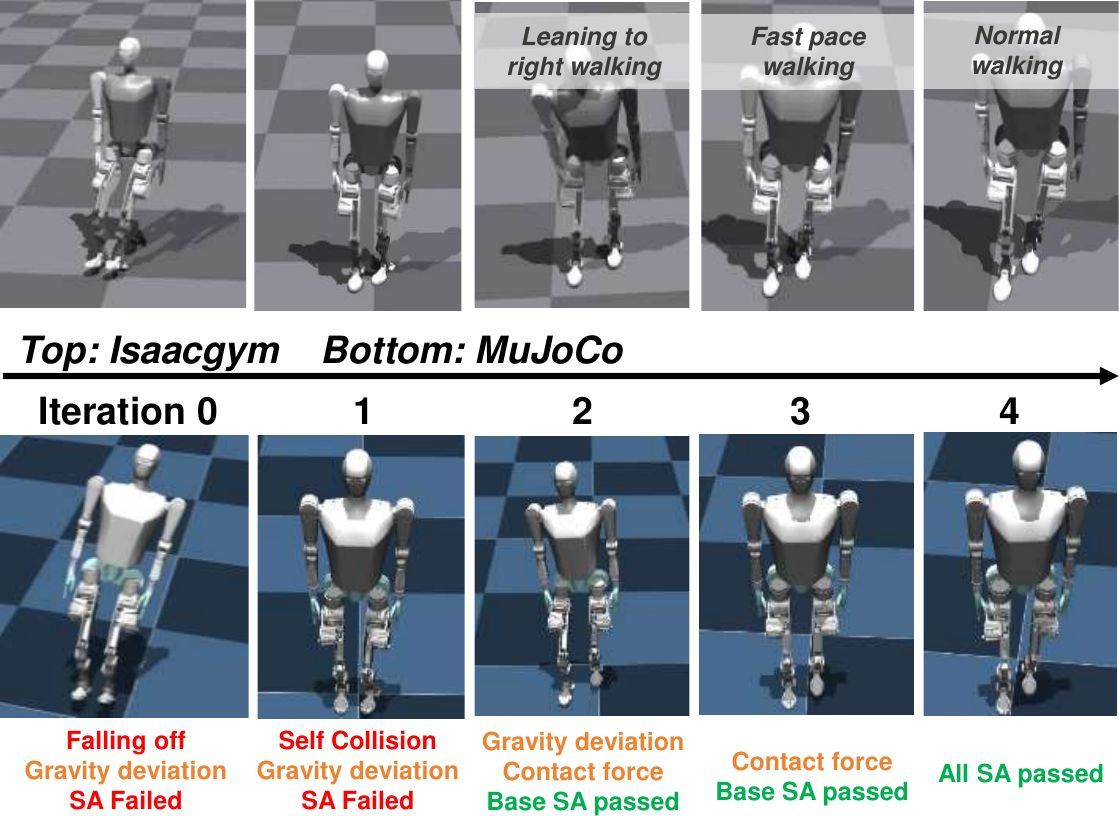}
        \caption*{H1 iterations.}
    \end{minipage}%
    \hspace{0.02\textwidth}
    \begin{minipage}{0.3\textwidth}
        \centering
        \includegraphics[height=3.5cm, keepaspectratio]{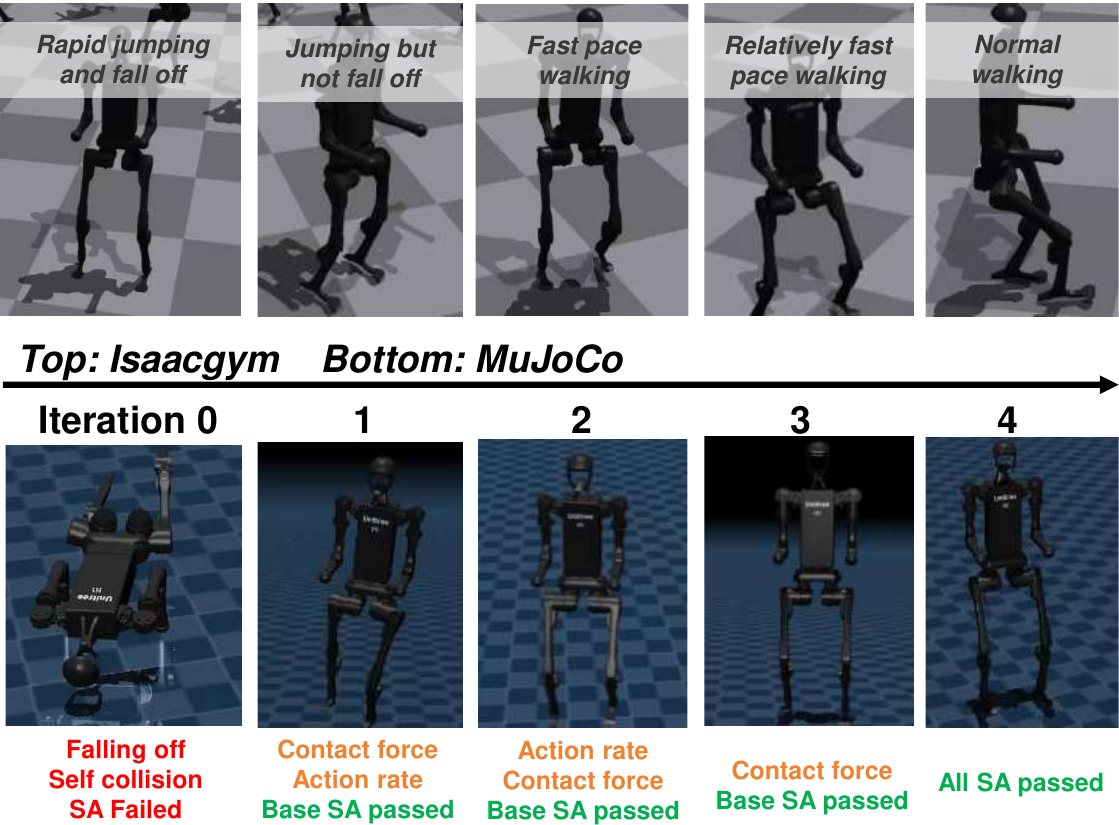}
        \caption*{Turin iterations.}
    \end{minipage}
    \caption{Best generation policies for 5 iterations evaluated in training and evaluating environments. For P1, it is evaluated also in reality. Figures indicate the policy improvement among iterations and safety check $SA$ feedback below each iteration. Colors indicate the severeness of different $SA$ indicators.}
    \label{fig:allexp}
\end{figure*}

We evaluated each LLM base by testing prompt design and various augmentation. The temperature for generation is set to \textbf{0.4} for all models. We define success rate (number of code that can be executed for all generation, SR) and max normalized reward success (max $R_{succ} / num\_steps\_per\_env$, the average on three experiments, Max S.) as two important indicators. The test result among the LLMs are shown as follows. We noticed that Deepseek-R1 and Claude-3.5-Sonnet shows best performance. 
% Success rates—defined as the proportion of samples processed successfully in the first iteration—were compared across four configurations: basic prompts, prompts with code references, safety regulations, and both. Fig. \ref{fig:success_rate} shows that adding code references and safety regulations improves LLM performance in generating usable reward functions. As gpt-4 and gpt-4o show similar performance with complete prompts, gpt-4o is recommended as the primary model.

\begin{table}[ht]
    \centering
    \begin{tabular}{cccccc}
        \toprule
        \textbf{Name}  & \textbf{P1} & \textbf{H1}   & \textbf{Turin} & \textbf{SR} & \textbf{Max S.} \\
        \midrule
        GPT-4o  & 228/240 & 173/240 & 52/240 & 62.9\% & 0.824 \\
        Claude-3.5 & 232/240 & 213/240 & 201/240 & 89.7\% & 1.032 \\
        Deepseek-R1 & -/- & 232/240 & 211/240 & 92.3\% & 1.060 \\

        \bottomrule
    \end{tabular}
    \caption{Number of run-able code / total generation on different bot tasks, success rate, max normalized reward success for different LLMs.}
    \label{tab:llmfn}
\end{table}

% \begin{figure}[htbp]
%     \centering
%     \includegraphics[width = 0.45\textwidth]{imgs/success_rate.png}
%     \caption{Success rate for LLM generated reward functions}
%     \label{fig:success_rate}
% \end{figure}

% We validated the effect of incorporating safety regulations into reward functions by comparing models with and without safety prompts in Isaac Gym and Gazebo environments. Fig. \ref{fig:safe_unsafe} shows the operational postures and IMU spatial angles of safe models (c)(d) versus unsafe models (a)(b). Results indicate that, even without real-world feedback, safety prompts effectively constrain robot behavior, supporting practical deployment.

% \begin{figure}[htbp]
%     \centering
%     \includegraphics[width = 0.39\textwidth]{imgs/crop_fig3.pdf}
%     \caption{Model behavior with and without safety regulation prompts}
%     \label{fig:safe_unsafe}
% \end{figure}

\begin{figure}[htbp]
    \centering
    \includegraphics[width = 0.49\textwidth]{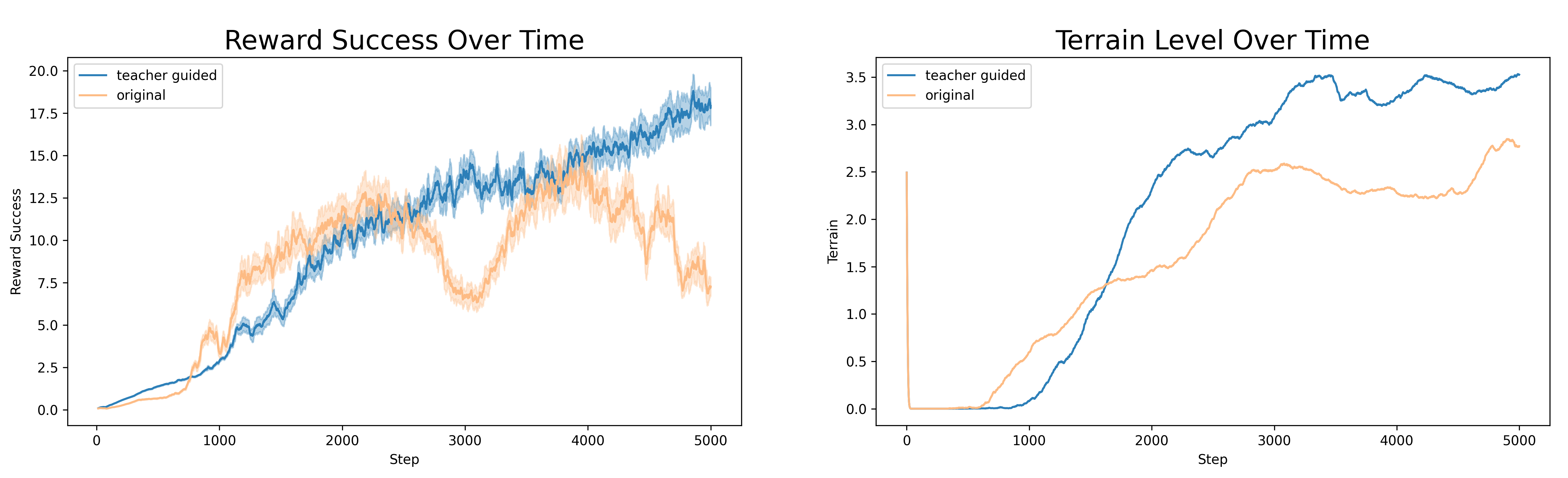}
    \caption{Reward success, terrain level for teacher guided and original RL training with human-engineered rewards, using P1 as example. Blue line shows trending of training with teacher, and orange one is pure PPO.}
    \label{fig:rew_success}
\end{figure}

To show the strength of reference policy, we take the training of P1 as example, set the reference policy using human-engineered reward function trained policy (5000 epochs training, flat terrain), and compare the training result using same reward function under trimesh terrain. The teacher coefficient is set to $\beta = 0.5$ with curriculum learning on. Performance was measured using \textit{reward success} and \textit{terrain level}. Results in Fig. \ref{fig:rew_success} showed that the teacher-guided model had more stable training, with faster and less volatile reward growth.
% we evaluated teacher-guided models under artificial reward functions, training each for 5,000 iterations on complex terrain. The original model used only an artificial reward function, while the teacher-guided model was instructed by an operational ONNX model with $\beta = 5.0$. Performance was measured using \textit{reward success} and \textit{terrain level}. Results showed that the teacher-guided model had more stable training, with faster and less volatile reward growth.

For the Homomorphic evaluation part, the following table shows some evaluation indicators before and after the conversion for P1 training, and the evaluation results under the best model. It can objectively reflect some differences from simulation to reality. Each column seperately represents reward function name, reward in isaac gym, homomorphic measurement in gazebo, in reality, and the mapped tracking result in reality (30 seconds of tracking, where env configuration requires at least 1.6s tracking).

\begin{table}[ht]
    \centering
    \begin{tabular}{ccccc}
        \toprule
        \textbf{Name}  & \textbf{Gym} & \textbf{Gazebo} & \textbf{Reality} & \textbf{Mapping (real/targ)}\\
        \midrule
        Track lin vel  & 0.8893 & 0.9138 & 0.8504 & 0.86 (1.0) m/s\\
        Track ang vel & 0.7652 & 0.6500 & 0.6013 & 0.06 (0.10) rad/s\\
        Feet distance & -0.31 & -0.00 & -0.00 & $>0.1$m ($>0.1$m) \\
        Standing still\textsuperscript{1} & -5.35 & -6.16 & - 11.20 &  5.8 (30) s \\
        Survival time & 0.86 & 1.0 & 1.0 & 30 (30) s\\

        \bottomrule
    \end{tabular}
    \footnotesize{Note: \textsuperscript{1}Standing still requires robot to maintain speed 0 for certain time.}
    \caption{Examples of homomorphic evaluation for pointfoot robot.}
    \label{tab:2}
\end{table}

For the safety check criterion $SA$, we define two levels: the \textbf{base} criterion, which includes falling off, self-collision (not implemented for P1\footnote{Contact detection extraction is implemented in MuJoCo only}), and violating DoF and torque limits; and the \textbf{warning} criterion, which includes high feet contact forces (not implemented for P1), high action rate, high total torque, and large gravity projection deviation. Policies passing the base safety tests are considered deployable, and those without warnings are regarded as safe. We conducted 3 training groups for each robot, generating 240 samples per robot ($5 \times 16$ results per run). Models passing the base test are labeled as positive, and those passing both tests as strict positive. The ground truth is established by manually evaluating each policy’s behavior in simulation to identify true positives and negatives. We then compute the average precision (AP) and average recall (AR) to assess $SA$ performance. We also list another success rate (SR) defined as the percentage of policy passed the safety check. Evaluation proved that properly implemented $SA$ can identify whether policy is safe under most scenarios. This also shows that prompts with better initial prompts leads to higher success rate.

\begin{table}[ht]
    \centering
    \begin{tabular}{cccccc}
        \toprule
        \textbf{Name}  & \textbf{AP} & \textbf{AR}   & \textbf{S-AP} & \textbf{S-AR} & \textbf{SR (S-SR)} \\
        \midrule
        P1 Pointfoot  & 94.3 & 96.4 & 100.0 & 92.6 & 58.7 (26.2)\\
        Unitree H1 & 100.0 & 97.5 & 93.3 & 84.0 & 65.4 (18.8) \\
        Turin Robot & 97.6 & 96.1 & 83.8 & 93.9 & 52.9 (15.4) \\

        \bottomrule
    \end{tabular}
    \caption{Precision and recall for safety criterion in percentage form, `s' label means strict indicator.}
    \label{tab:apar}
\end{table}

\subsection{Framework Analysis}

% The experimental setup involved first establishing a baseline using a manually designed reward function. We then trained our model from scratch in flat and complex terrain environments, sed to simulate the general situation of users first using the \textit{AnyBipe} framework. The robot was trained to track speed commands, avoid falls, and walk on various terrains using a simplified reward function with basic components like velocity tracking, action rate, and balance judgment, without scale-related information. Over five rounds ($N=16$ samples each), the best model from each round was tested in Gazebo and real-world scenarios. The basic training framework occupies 7G of GPU memory, and the corresponding training time is 79 hours. As shown in Fig. \ref{fig:exp3}, initial iterations had issues like excessive movement (Iteration 0) and unnatural joint postures (Iterations 1 and 2), but these were corrected by Iteration 5, achieving a natural gait.
The experimental setup is summarized in \autoref{tab:3}. For each type of experiment, we present results using different LLM bases\footnote{The choice of LLMs varies for different robots. P1 showed promising results with GPT-4o, and alternative models did not significantly improve performance. We intended to use Deepseek-R1, but its official API was unstable during the experiment (Jan. 2025 to Feb. 2025), so we opted for Claude-3.5-Sonnet instead.}. Detailed results and setup for similar experiments can be found on GitHub.
\begin{table}[ht]
    \centering
    \begin{tabular}{ccccc}
        \toprule
        \textbf{Name}  & \textbf{LLM} & \textbf{Train Iter} & \textbf{Sim Env} & \textbf{Real Env}\\
        \midrule
        P1 Pointfoot  & GPT-4o & 5,000 & Gazebo & $\checkmark$\\
        Unitree H1 & Claude-3.5 &2,000 & MuJoCo & $\times$\\
        Turin Robot & Deepseek-R1 & 2,000 & MuJoCo & $\times$ \\

        \bottomrule
    \end{tabular}
    \caption{Framework experimental set-up.}
    \label{tab:3}
\end{table}

\begin{table*}[htbp]
    \centering
    \begin{tabular}{ccccccc}
        \toprule
        \textbf{Name}  & \textbf{Iter 0} & \textbf{Iter 1} & \textbf{Iter 2} & \textbf{Iter 3} & \textbf{Iter 4} & \textbf{Human Engineered}\\
        \midrule
        P1 Pointfoot (iter=5000)  & 14.7 (5.5) & 7.1 (6.3) & 9.2 (8.7) & 15.5 (11.2) & 25.6 (20.8) & 19.2 (19.2) \\
        %P1 Pointfoot (iter=2500)  & 14.7 (5.5) & 7.1 (6.3) & 9.2 (8.7) & 15.5 (11.2) & 25.6 (20.8) & 19.2 (19.2) \\
        Unitree H1 (iter=2000) & 0.0097 (0.0035) & 21.2 (12.4) & 25.2 (21.7) & 25.0 (22.4) & 25.8 (24.9) & 23.7 (23.7)\\
        %Unitree H1 (iter=1000)& 0.0097 (0.0035) & 21.2 (12.4) & 25.2 (21.7) & 25.0 (22.4) & 25.8 (24.9) & 23.7 (23.7)\\
        Turin Robot (iter=2000)& 10.2 (1.96) & 52.3 (42.9) & 58.1 (51.8) & 66.3 (55.4) & 67.0 (57.8) & None (None)\\
        %Turin Robot (iter=1000)& 10.2 (1.96) & 52.3 (42.9) & 58.1 (51.8) & 66.3 (55.4) & 67.0 (57.8) & None (None)\\

        \bottomrule
    \end{tabular}
    \caption{Reward success $\mathfrak{E}_{train}$, over iterations compared to human engineered ones in format of batch max (batch average). The upper bound of the term equals $num\_step\_per\_env$. }
    \label{tab:rews}
\end{table*}

Fig. \ref{fig:allexp} describe the best training result iterating over framework iterations, along with $SA$ feedback indicating different levels of safety violation (\color{red}red \color{black} for not safe, \color{yellow}yellow \color{black} for potential problem warning, and \color{green}green \color{black} for passed check). P1 passes all $SA$ check and all reality deployments are listed. The other two are listed in training-simulation pairs. These experiments prove that \textit{AnyBipe} is capable of guiding LLM to realize what problems might occur in certain reward function implementation, and can act correspondingly to solve the problem. By iterations the unsafe problems are properly handled while maintaining code set effective.
% \begin{figure}[htbp]
%     \centering
%     \includegraphics[width=0.42\textwidth]{imgs/expp1.pdf}
%     \caption{Deployment results and SA feedback for P1.}
%     \label{fig:exp3}
% \end{figure}

\begin{figure}
    \centering
    \includegraphics[width = 0.49\textwidth]{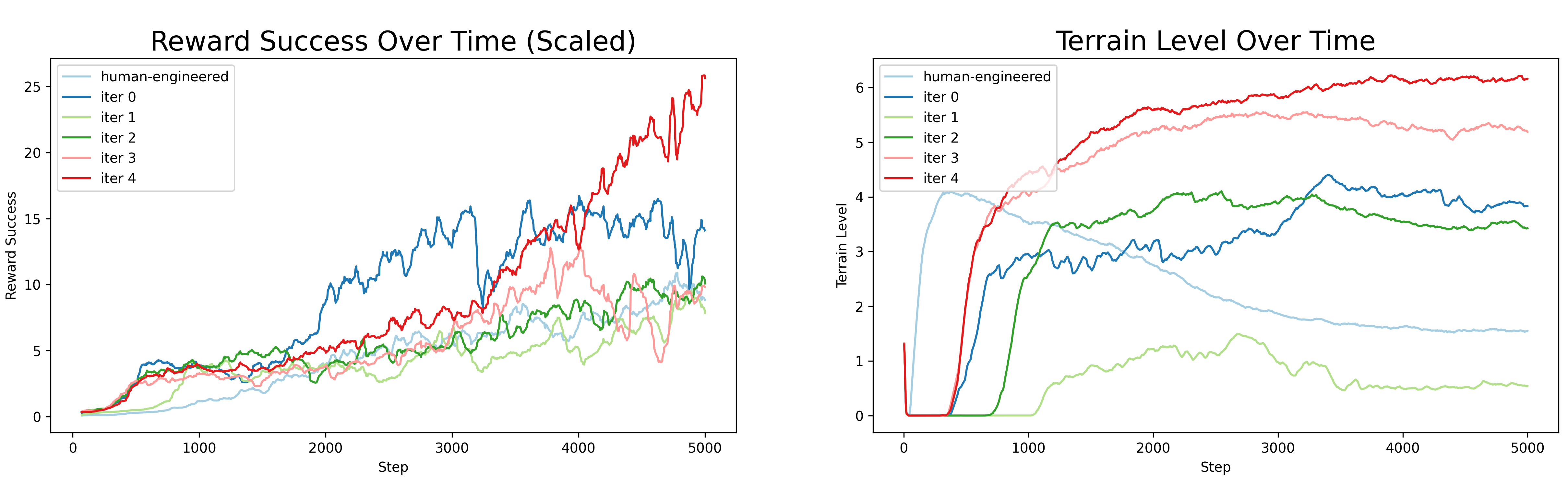}
    \caption{Two success indicators for P1 training, compared among iterations and with open-source human engineered rewards (light blue line), regarded as current state-of-the-art (SOTA) method. The result indicates LLMs can explore better forms or combinations for reward functions for given RL task.}
    \label{fig:rew_final}
\end{figure}

% To demonstrate the effectiveness of the LLMs' reward function improvements during training, we evaluate its performance using \textit{reward success} and \textit{terrain level} in tasks like velocity tracking, safe walking, and navigating complex terrains. After comparing five iterations with manually designed reward functions in Fig. \ref{fig:rew_final}, we found that the LLMs' reward function outperformed the manual version after just two iterations. Each subsequent iteration further enhanced training speed and performance, all achieved without human intervention.

\begin{figure}[htbp]
    \centering
    \includegraphics[width = 0.48\textwidth]{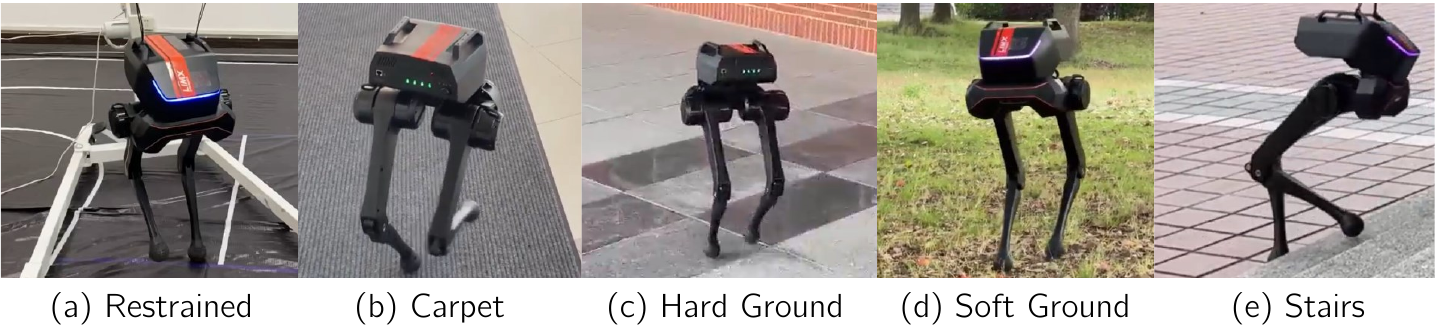}
    \caption{Reality experiments for P1 conducted on different terrains, adopting \textit{AnyBipe}'s best policy.}
    \label{fig:terrain}
\end{figure}

We visualize the reward and terrain level curves for P1 training, comparing them to human-engineered reward curves extracted from TensorBoard logs in Fig. \ref{fig:terrain}. For the three experiments, we report the average success rate at the midpoint and end of training for 5 iterations, comparing them with human-engineered metrics (considered state-of-the-art, as they are available open-source) in \autoref{tab:rews}. Our results demonstrate that \textit{AnyBipe} can identify more effective reward function combinations after several iterations without human intervention. The framework can also autonomously explore suitable code implementations in zero-shot scenarios when no reference is provided (as shown in the Turin task). This highlights the framework’s potential for training newly designed robot configurations.

Experiments in Fig. \ref{fig:terrain} indicates that the trained policy can be truly deployed in reality, and maneuvers different kinds of terrain types.

\section{Conclusion}
\textit{AnyBipe} presents an end-to-end framework for training and deploying bipedal robots, leveraging state-of-the-art Large Language Models (LLMs) to design reward functions tailored to specific tasks. The framework provides interfaces that allow users to supply coding references and integrate pre-existing models to facilitate the training process. Furthermore, it incorporates feedback from both simulated and real-world test results, enabling the execution of Sim-to-Real tasks without human supervision while also offering improvement directions to the LLMs. We have validated the effectiveness of each module, as well as the system's capacity to guide the robot in learning locomotion in complex environments, progressively improving the model by either creating new reward functions from scratch or refining existing ones. Moreover, this framework demonstrates potential for transferability to other robotic task planning scenarios.

Despite these strengths, there remain areas for further improvement. Treating observations and reward terms solely as human factors may not fully capture human preferences, and visual factors should be considered as well. Also, the framework automates the whole training process, but still requires human to implement the basic RL environment following legged gym and our template, not generating environment directly from LLM. Additionally, incorporating Vision-Language Models (VLMs) into the safety criterion could improve the precision of judgment.

Future work will focus on advancing the framework in key areas: first, expanding its application to a broader range of robotic tasks to assess its generalizability, testing its effectiveness beyond locomotion; second, introducing a supervisory process during the training stage to allow the framework to autonomously determine the duration of reinforcement learning training, rather than relying on a fixed total length; and third, refining the model evaluation process by incorporating visual and conceptual feedback to achieve a more comprehensive state estimation.

%%%%%%%%%%%%%%%%%%%%%%%%%%%%%%%%%%%%%%%%%%%%%%%%%%%%%%%%%%%%%%%%%%%%%%%%%%%%%%%%

%%%%%%%%%%%%%%%%%%%%%%%%%%%%%%%%%%%%%%%%%%%%%%%%%%%%%%%%%%%%%%%%%%%%%%%%%%%%%%%%



% \section*{ACKNOWLEDGMENT}

% The preferred spelling of the word ÒacknowledgmentÓ in America is without an ÒeÓ after the ÒgÓ. Avoid the stilted expression, ÒOne of us (R. B. G.) thanks . . .Ó  Instead, try ÒR. B. G. thanksÓ. Put sponsor acknowledgments in the unnumbered footnote on the first page.

%%%%%%%%%%%%%%%%%%%%%%%%%%%%%%%%%%%%%%%%%%%%%%%%%%%%%%%%%%%%%%%%%%%%%%%%%%%%%%%%

% References are important to the reader; therefore, each citation must be complete and correct. If at all possible, references should be commonly available publications.

\newpage

\bibliographystyle{IEEEtran}
\small\bibliography{reference}
\balance
\end{document}